\documentclass[runningheads]{llncs}
\usepackage{graphicx}
\usepackage{hyperref}

\usepackage{amsmath}
\usepackage{algorithm, algorithmicx}
\usepackage{algpseudocode}
\usepackage{float}
\usepackage{caption}
\usepackage{subcaption}
\usepackage{setspace}
\usepackage{booktabs}
\begin{document}
\title{Joint Chinese Word Segmentation and Span-based Constituency Parsing}
%
\author{Zhicheng Wang, Tianyu Shi, Cong Liu\thanks{Cong Liu is the corresponding author.}}

\institute{School of Computer Science and Engineering\\
Sun Yat-sen University\\
Guangzhou, China\\
\email{\{wangzhch23, shity3\}@mail2.sysu.edu.cn, liucong3@mail.sysu.edu.cn}}
\maketitle              
\begin{abstract}
In constituency parsing, span-based decoding is an important direction. However, for Chinese sentences, because of their linguistic characteristics, it is necessary to utilize other models to perform word segmentation first, which introduces a series of uncertainties and generally leads to errors in the computation of the constituency tree afterward. This work proposes a method for joint Chinese word segmentation and Span-based Constituency Parsing by adding extra labels to individual Chinese characters on the parse trees. Through experiments, the proposed algorithm outperforms the recent models for joint segmentation and constituency parsing on CTB 5.1.
\keywords{Constituency parsing \and Chinese Word Segmentation}
\end{abstract}

\section{Introduction}
In natural language processing, constituent parsing is fundamental, which recognizes the phrase structure and syntactic tree of a sentence. Constituency parsing is useful for a variety of upstream tasks, such as translation and sentiment analysis. A neural parser often consists of an encoding module and a decoding module. The encoding module obtains the context representation of each word in a sentence. With the rapid development of representation learning, the encoding model has gradually evolved from the LSTM to Transformer~\cite{vaswani2017attention} with stronger representation capability. In terms of decoding, there are also many different types of decoding algorithms, such as transition-based decoding~\cite{watanabe2015transition,liu2017order}, 
span-based decoding~\cite{stern-etal-2017-minimal,kitaev2018constituency}, and sequential-to-sequence decoding~\cite{shen2018straight,gomez2018constituent}.

Much of the previous work has focused on improving encoders, e.g. using more semantic and contextual information to improve performance~\cite{mrini2019rethinking}. In the decoding stage, a span-based decoder is popular. Stern et al.~\cite{stern-etal-2017-minimal} scores labels and splits separately, and calculates the parse tree with the highest score from the bottom up through a dynamic programming algorithm. In addition, they provide a computationally efficient greedy top-down inference algorithm based on recursive partitioning of the input. 

However, in contrast to English, where punctuation marks and spaces between words serve as natural constituency markers, Chinese sentences must first undergo word segmentation using alternative models or algorithms due to differences in linguistic characteristics from those of English. A number of uncertainties are introduced when additional models or methods are used. For instance, incorrect word segmentation will typically result in problems when computing the constituency tree afterward.

In this work, we attempted to parse at the character level to avoid relying on external segmentation tools, which results in joint Chinese word segmentation and constituency parsing. In prior work, Xin et al.~\cite{xin2021n} extends the label set with POS tags for n-ary tree parsing models. We take one step further and apply it to binary tree parsing. We first label each individual Chinese character with the label "@1" and then transform the tree into Chomsky normal form (CNF). We further use the label "@2" to denote nodes that are generated when binarizing the subtree for each word with more than two characters. 

Our approach surpasses a number of joint-task models for span-based parsing on the Chinese Penn Treebank. For joint tasks, the F1-measure of Chinese word segmentation and constituency parsing of our decoder are 99.05 and 91.94.

\section{Related Work}
\subsection{Early Models for Span-based Decoding}
Early constituency parsing methods are mainly based on grammar and statistics, such as probabilistic context-free grammar (PCFG). On this basis, the widely used CKY decoding algorithm is produced, which is essentially a dynamic programming algorithm. Then, Collins~\cite{collins1997three} extends the probabilistic context-free grammars and proposes a generative, lexicalised, probabilistic parsing model. After that, Matsuzaki et al.~\cite{matsuzaki2005probabilistic} defines a generative probabilistic model of parse trees with latent non-terminal splitting annotations. For a long time, such generative models have dominated constituency parsing.

In recent years, span-based parsers were presented, which used log-linear or neural scoring potentials to parameterize a tree-structured dynamic program for maximization or marginalization~\cite{finkel2008efficient,durrett2015neural}. As one of the most influential works, Stern et al.~\cite{stern-etal-2017-minimal} presents a minimal neural model for constituency parsing based on independent scoring of labels and spans, and achieves state-of-the-art performance. Kitaev and Klein~\cite{kitaev2018constituency} further improves the encoder with factored self-attention, which disentangles the propagation of contextual information and positional information in the transformer layers. On this basis, Mrini et al.~\cite{mrini2019rethinking} proposes the Label Attention Layer, which uses extra labels to encode task-related knowledge similar to the more recently proposed prompting technique~\cite{liu2021pre}.

The proposed parser adopts the encoder from previous work~\cite{mrini2019rethinking}.

\subsection{Joint Chinese Word Segmentation and Constituency Parsing}
Unlike English, Chinese sentences consist of single characters without segmentation. For Chinese constituency parsing, sentences are first segmented using external segmentatoare before inputting into the model~\cite{wang2006fast}. Prior work attempts to combine the two tasks. Qian and Liu~\cite{qian2012joint} trains the two models separately and incorporates them during decoding. Zhang et al.~\cite{zhang2013chinese} extends the notion of phrase-structure trees by annotating the internal structures of words. They label each individual Chinese character and add structural information to the label to improve parsing performance. Xin et al.~\cite{xin2021n} extends the label set with the POS tags for n-ary tree parsing models.

In this work, we label "@1" on each individual Chinese character and "@2" on sub-words, We can therefore label word segmentation and label spans for parsing at the same time.

\section{Model}

\subsection{Preliminaries}

\begin{figure}
\begin{subfigure}{.22\linewidth}
\centering
\includegraphics[scale=0.14]{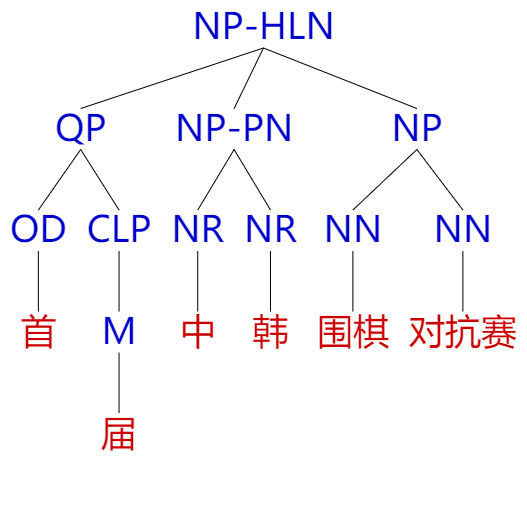}
\caption{Original Tree}
\label{fig:tree1}
\end{subfigure}%
\begin{subfigure}{.31\linewidth}
\centering
\includegraphics[scale=0.14]{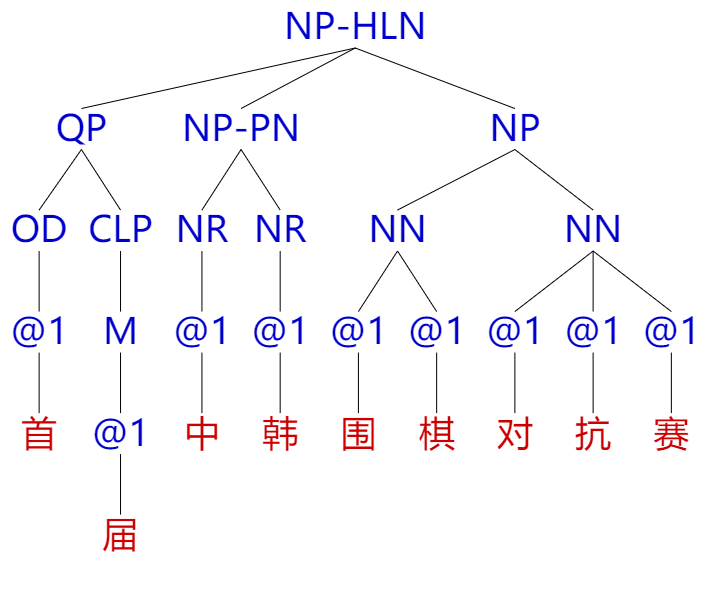}
\caption{Add Extra Label}
\label{fig:tree2}
\end{subfigure}
\quad
\begin{subfigure}{.44\linewidth}
\centering
\includegraphics[scale=0.14]{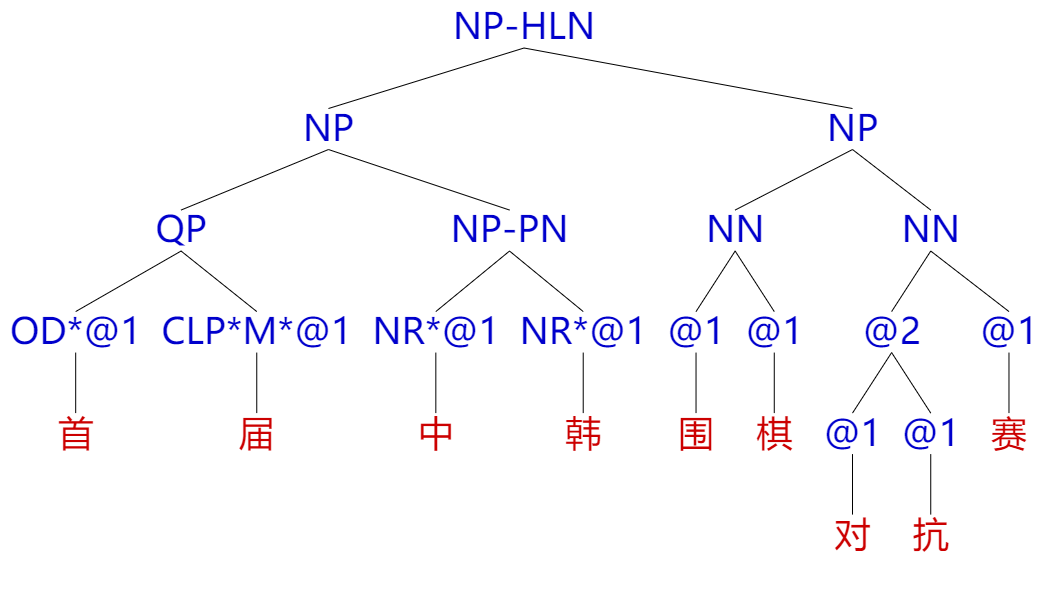}
\caption{Left Binarization}
\label{fig:tree3}
\end{subfigure}
\caption{Illustration of how we convert a parse tree into a binarized character-level tree.}
\label{fig:tree}
\end{figure}

An example parse tree is shown in Figure~\ref{fig:tree1}, which is the model input. We start by adding extra labels "@1" to each individual Chinese character, as shown in Figure~\ref{fig:tree2}. Individual Chinese characters are treated as words and words as phrases. In order to decode in a CKY-liked algorithm, we transform the tree into Chomsky Normal Form (CNF), as shown in Figure~\ref{fig:tree3}. Following previous work~\cite{zhang2020fast}, consecutive unary productions are merged into a single node. For words with more than two Chinese characters, we introduced another label "@2" during binarizing. The CNF tree is converted back into the n-ary tree after decoding.

\subsection{Encoding}
A sentence is denoted by $x = \{x_1,x_2,...,x_n\}$, where $x_i$ is the $i^{th}$ word, and $n$ is the sentence length. 

In the encoding stage, our goal is to obtain the score of each span $(i,j)$, which will be used in the decoding stage to obtain the best parse tree. In the embedding layer, we use the pretrained language model Chinese BERT to generate the contextual embeddings. We denote $e_i$ the embedding vector of the $i^{th}$ word in the sentence.
\begin{equation}
e = BERT(x), e = \{e_1,e_2,...,e_n\}
\end{equation}

In the encoding layer, the Transformer~\cite{vaswani2017attention} is selected for extracting the contextual features, denoted by $g_i$. Following previous work, we also stack another layer of Label Attention Layer (LAL)~\cite{mrini2019rethinking} on top of the Transformer layers. The final contextual embedding is denoted by $h_i$, whose dimension is 1024.
\begin{equation}
g = Transformer(e)
\end{equation}
\begin{equation}
h = LAL(g)
\end{equation}

In the scoring phase, we convert the word representation into span representation, denoted by $H(i,j)$. We use a two-layer MLP to calculate the scores of span $(i,j)$ for different labels from $H(i,j)$.
\begin{equation}
s_{span}(i,j,\cdot) = MLP(H(i,j))
\end{equation}

where $s_{span}(i,j,l)$ denotes the score of assigning label $l$ to span $(i,j)$.
\subsection{Decoding}
\begin{figure}
\begin{subfigure}{.32\linewidth}
\centering
\includegraphics[scale=0.17]{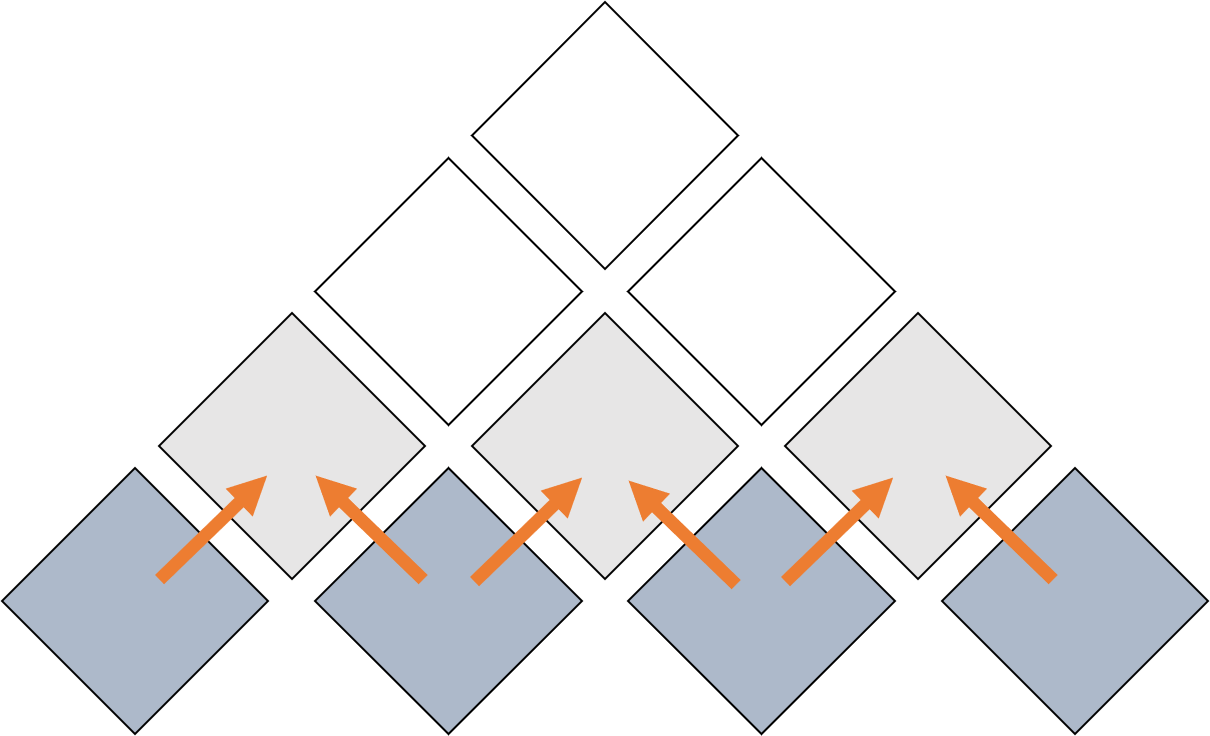}
\caption{}
\label{fig:rule1}
\end{subfigure}
\begin{subfigure}{.32\linewidth}
\centering
\includegraphics[scale=0.17]{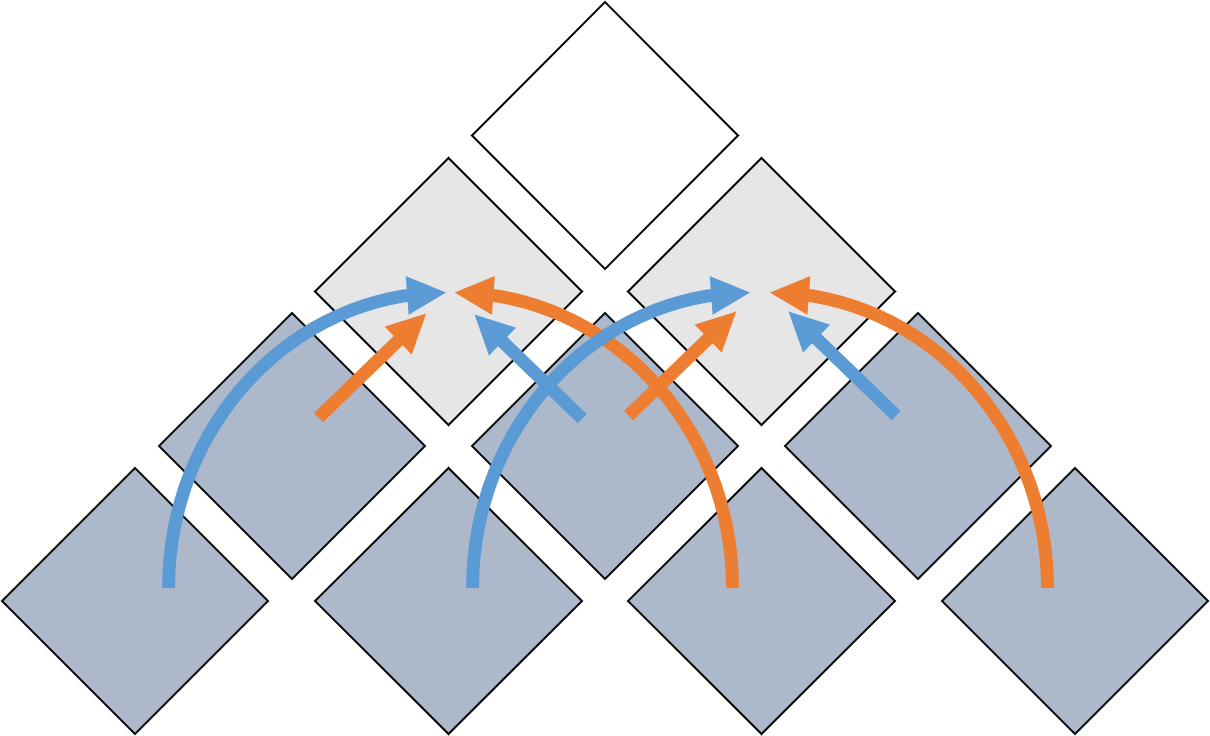}
\caption{}
\label{fig:rule2}
\end{subfigure}
\begin{subfigure}{.32\linewidth}
\centering
\includegraphics[scale=0.17]{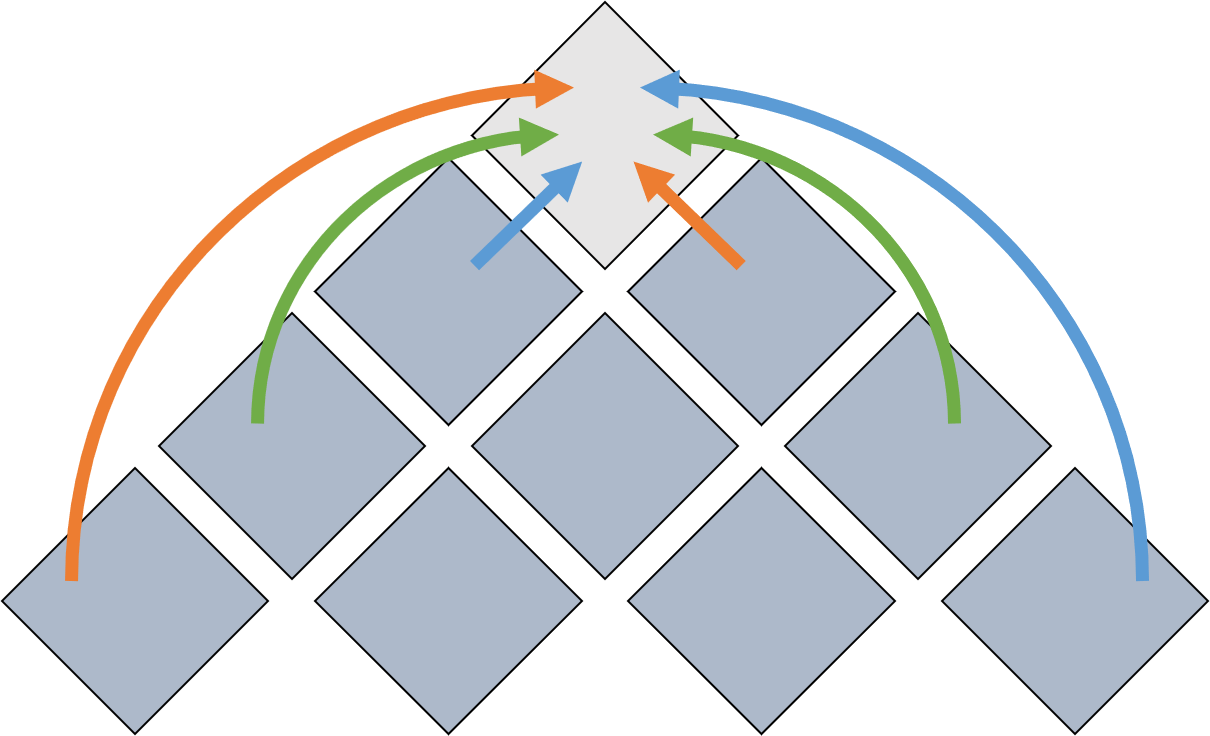}
\caption{}
\label{fig:rule3}
\end{subfigure}
\caption{A bottom-up dynamic programming calculation. Different colored lines represent different splits. The best splits are recorded during the calculation.}
\label{fig:rule}
\end{figure}

Let $s_{tree}(i,j,l)$ denote the best score of span $(i,j)$ for the label $l$, which is calculated with a CKY bottom-up dynamic programming algorithm. In the base case of this algorithm, each span $(i,j)$ contains a single word, and $s(i,j,l)$ is given by the encoder, since the tree span $(i,j)$ contains a single node.
\begin{equation}
s_{tree}(i,j,l)=s_{span}(i,j,l)
\end{equation}

For a span $(i,j)$ covering more than one word, let $k (i<k<j)$ be a split point between its left and right children, $l$ be the label of the span, $l_1$ and $l_2$ be the labels of the left and the right children. The tree score is calculated by summing its subtree scores of the constituent tree and the optimal parse tree is the parse tree with the highest tree score:

\begin{equation}
s_{tree}(i,j,l) = \mathop{max}\limits_{i<k<j}\ [s_{span}(i,j,l) + s_{tree}(i,k,l_1) + s_{tree}(k,j,l_2)]
\end{equation}

During decoding, we generate the optimal tree on each span for each label in a top-down order. For the root span, we obtain the label $l_{best}$ through $\mathop{max} s_{tree}(i,j,l)$. Then, we can trace back the optimal split applied on each node to construct the optimal tree in a top-down order. The whole process is shown in Figure~\ref{fig:rule}.

\subsection{Training Loss}
We use two kinds of training losses. In the first 10 epochs, label loss is used to make the model converge quickly. Its calculation does not require to decode the optimal tree, and it is simply the sum of cross-entropy between the distribution of each span and its ground-truth label.
\begin{equation}
Loss_{label} = \sum_{(i,j)\ in\ T} CrossEntropyLoss(s_{span}(i,j),l_{i,j})
\end{equation}

For the rest epochs, we use tree loss, which is defined as the hinge-loss between the sum of the label scores of the spans in the predicted tree and that in the optimal tree.
\begin{equation}
Loss_{tree} = \sum_{(i,j)\ in\ T_{pred}} s_{span}(i,j,l_{i,j}) + 1 \\
- \sum_{(i,j)\ in\ T_{gold}} s_{span}(i,j,l_{i,j})
\end{equation}

\section{Experiments}
\subsection{Experimental Setup}
We evaluate our model for constituency parsing on Chinese Treebank 5.1~\cite{xue2005penn}. It consists of 17,544/352/348 examples in its training/validation/testing splits respectively. Each example is a parse tree with internal nodes associated with labels, and words associated with tags. We follow the previous work~\cite{liu2017order} and adopt the same method of pre-processing. POS tags are removed and not use as input features in both training and testing processes on CTB 5.1, following the previous work in Zhang et al.~\cite{zhang2020fast}. We employ the EVALB tool to calculate standard precision, recall and F1-measure as evaluation metrics. Models are trained solely on training data to be evaluate in the test set (not including validation data). Table. \ref{table:hyper} lists the hyper-parameters used in the implementation. The hyper-parameter settings for transformer and Label Attention Layer are the same with the previous work~\cite{mrini2019rethinking} and are therefore no longer listed. On a single GTX TITAN, the model is implemented in PyTorch. We employ the pre-trained Chinese BERT to compare with previous works.

\begin{table}
\centering
\caption{Hyper-parameters.}
\begin{tabular}{lc}
\hline
\textbf{Parameter} & \textbf{Value}\\
\hline
learning rate & $10^{-5}$ \\
decay factor & 0.5 \\
max decay & 10 \\
decay patience & 3 \\
\hline
\end{tabular}
\begin{tabular}{lc}
\hline
\textbf{Parameter} & \textbf{Value}\\
\hline
batch size & 250 \\
MLP layer & 2 \\
MLP hidden & 250 \\
dropout & 0.2 \\
\hline
\end{tabular}
\label{table:hyper}
\end{table}

\subsection{Performance}
Table. \ref{table:joint} indicates the overall performance of the joint task on the test set. We use the work in Xin et al.~\cite{xin2021n} as the baseline, which also uses Chinese BERT as the pre-trained model. It can be observed that the F1 measurements of Chinese word segmentation and Constituency Parsing of our proposed framework are 99.05 and 91.94, which outperform the baseline model.

\begin{table}
\centering
\caption{Joint-task performance on test set of CTB 5.1.}
\setlength{\tabcolsep}{5mm}{
\begin{tabular}{lcc}
\toprule[1.5pt]
\textbf{Model} & \textbf{Seg-F1} & \textbf{Par-F1}\\
\midrule
Qian and Liu~\cite{qian2012joint} & 97.96 & 82.85 \\
Wang et al.~\cite{wang2013lattice} & 97.86 & 83.42 \\
Zhang et al.~\cite{zhang2013chinese} & 97.84 & 84.43 \\
Xin et al.~\cite{xin2021n} & 98.92 & 91.84 \\
\midrule 
Ours & \textbf{99.05} & \textbf{91.94} \\
\bottomrule[1.5pt]
\end{tabular}}
\label{table:joint}
\end{table}

\subsection{Speed Analysis}
We use the test set of CTB5.1 to measure parsing speed. To reduce randomness, we conduct 10 experiments and averaged them. The results are shown in Table. \ref{table:speed}. The average processing speed is 50 sentences per second with a single RTX 3090. We compare with Xin et al.~\cite{xin2021n} as the baseline, which is also processed on a single RTX 3090. It can be seen that our model has better performance, and the speed of our model is 2.5 times the baseline.

\begin{table}
\centering
\caption{Speed comparison on CTB5.1 test set.}
\setlength{\tabcolsep}{5mm}{
\begin{tabular}{lc}
\hline
 & \textbf{Sents/sec}\\
\hline
Xin et al.~\cite{xin2021n} & 20\\
ours & 50\\
\hline
\end{tabular}}
\label{table:speed}
\end{table}

\section{Conclusion}
In this paper, we add extra labels for individual Chinese characters on the parse trees for joint Chinese word segmentation and constituency parsing. Experiments on CTB 5.1 yield several promising results. The proposed framework performs better than earlier work, with F1 improvements of 0.13 and 0.10 percent for the joint tasks of word segmentation and constituent parsing. Additionally, our model's computational speed is faster than earlier work.

\bibliographystyle{splncs04}
\bibliography{main}

\end{document}